# Deep Maxout Network-based Feature Fusion and Political Tangent Search Optimizer enabled Transfer Learning for Thalassemia Detection

Hemn Barzan Abdalla[1]*, Awder Ahmed[2], Guoquan Li[3], Nasser Mustafa[4], Abdur Rashid Sangi[5]

[1,4,5] Department of Computer Science, Wenzhou-Kean University, Wenzhou, China

[2] Department of Communication Engineering, Sulaimani Polytechnic University, Iraq

[3] School of Information and Communication Engineering, Chongqing University of Posts and Telecommunications, Chongqing, China

* Corresponding authors: habdalla@kean.edu

## Abstract

**Thalassemia is a heritable blood disorder which is the outcome of a genetic defect causing lack of production of hemoglobin polypeptide chains. However, there is less understanding of the precise frequency as well as sharing in these areas. Knowing about the frequency of thalassemia occurrence and dependable mutations is thus a significant step in preventing, controlling and treatment planning. Here, Political Tangent Search Optimizer based Transfer Learning (PTSO_TL) is introduced for thalassemia detection. Initially, input data obtained from a particular dataset is normalized in the data normalization stage. Quantile normalization is utilized in the data normalization stage, and the data are then passed to the feature fusion phase, in which Weighted Euclidean Distance with Deep Maxout Network (DMN) is utilized. Thereafter, data augmentation is performed, using oversampling method to increase data dimensionality. Lastly, thalassemia detection is carried out by TL, wherein a convolutional neural network (CNN) is utilized with hyperparameters from a trained model such as Xception. TL is tuned by PTSO and the training algorithm PTSO is presented by merging of Political Optimizer (PO) and Tangent Search Algorithm (TSA). Furthermore, PTSO_TL obtained maximal precision, recall and f-measure values of about 94.3%, 96.1% and 95.2%, respectively.**

## Introduction

Thalassemia is one of the most common types of genetic impairments worldwide. It is also described as inherited autosomal recessive disease that particularly affects people in Southeast Asia [2]. Each year, many babies are born with this thalassemia disease globally. It is referred to as acute hereditary anemia and also as the most frequent hemoglobin disorder [14]. "Thalassemia" is a word that is derived from two Greek words, namely "Thalassa" and "Haema" meaning "sea" and "blood". It was thus named owing to its higher occurrence in the Mediterranean nations [12]. Classically, there is higher occurrence of thalassemia in Middle Eastern, Southeast Asian and Mediterranean compared to North American and Northern European populations [1]. As it is a heritable disorder, the occurrence is owing to mutations in Deoxyribo Nucleic Acid (DNA) of the cells caused by the body's inadequate generation of Hemoglobin. Hemoglobin is the protein which permits Red Blood Cells (RBCs) to carry oxygen. An insufficiency of Hemoglobin lessens RBCs' survival rate, resulting in a lower count of RBCs flowing in the bloodstream, leading to a low oxygen supply in the body that can be dangerous [3]. There are several degrees of hypochromic microcytic anemia, depending on genetic faults of the alpha or beta globulin genes [26] [1]. Moreover, in this advanced global community and age of human migration, the occurrence of thalassemia is increasing in areas classically believed to have low occurrence, while screening and prevention programs may decrease the numbers of affected people in high occurrence areas [1].

Thalassemia causes the body to have decreased levels of Hemoglobin when compared to the normal level. Furthermore, it is caused by damaged or missing genes, which are accountable for the production of Hemoglobin [13]. It is mostly classified as alpha-thalassemia or beta-thalassemia, which depends on imperfect globin chains as well as fundamental molecular defects. It is a recessive gene disorder, thus medically applicable phenotypes arise due to homozygosity or heterozygosity and produce diverse globin gene defects [9]. Alpha-thalassemia has a complicated genetic heritage as a disorder that involves hemoglobin A1 as well as hemoglobin A2 genes. Each individual has four operational Alpha-globin genes, and two Alpha-globin alleles are inherited from each parent. Deletion of one or both of the Alpha-genes on chromosome 16 happens in about 95% of persons affected by Alpha-globin chain reduction [10]. Beta-thalassemia is naturally highly heterogeneous with regard to the molecular base of thalassemia disease. This kind of disorder is caused by widespread deletions happening inside the beta-globin gene on chromosome 11. A point mutation leads to the presence of stop codons in beta-globin mRNA. Other diverse mutations affect splicing sites and occasionally newer splicing sites develop [11]. Several tests are applied for diagnosing thalassemia, including Hemoglobin electrophoresis with A2 and F-quantitation, Chorionic Villus Sampling (CVS), Amniocentesis, Complete Blood Count (CBC) and Free Erythrocyte Protoporphyrin (FEP). However, these techniques are costly, prone to human mistakes, and consume more time. Furthermore, the recorded examinations are normally only executed in hospitals and clinics [13].

Machine Learning (ML) and Artificial Intelligence (AI) approaches connected with clinical imaging can be utilized for resolving the above-mentioned issues and enhancing medical diagnosis [13] [33]. These techniques are able to help the probers as well as doctors to make clinical decisions, and can address significant associated health care problems. There are four kinds of ML-based systems, namely unsupervised learning, reinforcement learning, supervised learning, and semi-supervised learning [25] [12] [32]. The most normally utilized learning approaches are logistic regression (LR), linear discriminant analysis (LDA), k nearest neighbor algorithm (KNN), Multilayer perceptron, Support Vector Machines (SVM), naïve Bayes (NB), Neural Networks (NN), Convolutional Neural Network (CNN) and decision trees (DT) [12] [30] [31]. While Deep Learning and AI are presently being utilized, an integrated approach of CNN-SVM can exceed the performance of feed-forward backpropagation NN, and can be combined with optical imaging for detecting thalassemia disease automatically [13] [36]. In recent years, Deep learning-enabled methods have enhanced classical techniques in several regions for better performances, particularly in computer vision. Deep learning utilizes numerous layered structures for performing non-linear information processing of various traditional deep neural network systems (DNN) [24] [15] [34], such as VGGNet, Inception-ResNet, SENet, ResNet, and AlexNet. The outcomes of these deep learning methods display entire learning features and inherent state details of data with higher efficiency than artificial features. Currently, deep learning approaches are one of the newer research trends in AI [15] [32].

The primary intention of this research is to introduce a new approach called PTSO_TL for the detection of thalassemia. The thalassemia syndromes consist of diverse levels of hemoglobin disorders owing to deficient or reduced generation of actual globin chains. Thalassemia is the most general recessive disease across the globe. Here, input data is taken from a particular database to perform thalassemia detection. Then, input data is subjected to data normalization, wherein redundancy is eliminated utilizing quantile normalization. Thereafter, feature fusion is performed in normalized data utilizing Weighted Euclidean Distance with DMN. Afterward, data augmentation is carried out in fused data utilizing the oversampling method. Lastly, thalassemia detection is conducted by CNN with TL, wherein CNN is utilized with hyperparameters from trained models such as Xception. The CNN with TL is trained to utilize PTSO, which is an amalgamation of PO and TSA. From the thalassemia detection phase, detection output is achieved.

The foremost contribution is elucidated as follows:

❖ **Proposed PTSO_TL for thalassemia detection:** For thalassemia detection, PTSO_TL is designed in this work. Here, thalassemia detection is performed utilizing CNN with TL, where CNN is employed with hyperparameters from trained models such as Xception. The training of CNN with TL is done by PTSO, which is an incorporation of PO and TSA.

The below sections are arranged as follows: section 2 contains the literature survey, section 3 explains the PTSO_TL methodology, section 4 presents the s PTSO_TL assessment results, and section 5 draws conclusions on PTSO_TL.

# Motivation

Thalassemia is an inherited genetic disease caused by alterations of beta or alpha-globin genes that result in the irregular synthesis of Hemoglobin. Many techniques have been designed for the detection of thalassemia, but those methods are time-consuming, costly, and labor-intensive. This motivated research to devise a technique for thalassemia detection by reviewing existing methods. The methods applied in earlier research for detection of thalassemia are explicated in this section.

## Literature Survey

Fu *et al*. [1] developed SVM to differentiate thalassemia from non-thalassemia utilizing simpler variables, but failed to validate this approach using many huge databases. Phirom *et al*. [2] designed the DeepThal method for predicting thalassemia, which was highly helpful in identifying risk as well as prenatal diagnosing of thalassemia quickly. However, they failed to provide user-friendly applications for ease of use in everyday practice. Rustam *et al*. [3] introduced supervised ML for the prediction of thalassemia. In this method, Synthetic Minority Oversampling Technique (SMOTE) and Adaptive Synthetic (ADASYN) were utilized to resolve dataset imbalance issues to reduce the bias of the model against major classes. Laengsri *et al*. [4] presented ThalPred for differentiating thalassemia, which achieved better discrimination, and overfitting issues were also lessened, but in this approach, a single parameter was not sufficient to differentiate between two criteria.

Doan *et al*. [5] introduced ML for detecting maternal carriers of general thalassemia and provided better estimation by decreasing the diverse forms load of α-thalassemia that is caused by the deletion mutations. The data utilized for training this model were based on Multiplex gap polymerase chain reaction (GAP-

PCR) and thus model performance was limited by the inherence accuracy of the GAP-PCR. Ou *et al*. [6] devised a spent culture medium for testing b-thalassemia that was optimized for obtaining higher concentrations of Deoxyribonucleic acid (DNA) without affecting the embryo development, although achieving a higher grade of fragmentation would have improved the performance of the model. Wang *et al*. [7] developed Loop-Mediated Isothermal Amplification for quick diagnosis of the thalassemia genes. In this technique, screening of genetic mutations and deletions was done rapidly, but it required re-designed primer sets to detect mutations with maximal specificity. Shuang *et al*. [8] designed a Logarithm-based XS-1 formula for distinguishing thalassemia. This method diagnosed genetic assessment, which offered prediction at an earlier stage, but did not consider the source or the computation scheme of the hematological parameters, which would have attained much better results.

### Challenges

The existing methods' challenges are interpreted as follows:

- ➢ SVM designed in [1] for discriminating thalassemia, as well as non-thalassemia, was not applicable to other populations, although diverse cohorts or populations exist, and the machine learning system would need to be modified or else re-established for applying to another cohort.
- ➢ In [3], the supervised ML developed for predicting thalassemia did not increase the size of the dataset for enhancing model performance and attaining the best classification accuracy in terms of β-Thalassemia.
- ➢ The method called ThalPred in [4] was devised for differentiating thalassemia by supplementing current indices to help health-care providers, but it did not consider partitioning of patients' ages to train the dataset and make the system more specific.
- ➢ The introduced technique ML in [5] was simple to scale up, but did not gather the ethnicity details of participants, and thus the prevalence record failed to reflect the occurrence among certain ethnic groups.
- ➢ Thalassemia is generally caused by sequence variants in Hemoglobin subunit Beta or Hemoglobin A1, and Hemoglobin A2 genes. The range of diverse approaches, such as Sanger sequencing, reverse hybridization, and GAP-PCR, are classically needed for accessing every variant. However, a long turn-around time was necessary to detect alpha and beta thalassemia sequence variation in patients.

# Proposed PTSO_TL for thalassemia detection

Thalassemia is a varied group of heritable hemoglobin disorders, with beta-thalassemia being the most widespread across the globe. In this research, PTSO_TL is introduced for the detection of thalassemia. Input data drawn from the database are normalized utilizing quantile normalization in the data normalization phase. Then, Weighted Euclidean Distance with DMN is utilized to carry out feature fusion in normalized data. Thereafter, the oversampling method is employed to execute data augmentation. Finally, the detection of thalassemia is performed by CNN with TL, in which CNN is utilized with hyperparameters from trained models such as Xception. PTSO is employed to train CNN with TL; however, PTSO is designed by integrating PO and TSA. Figure 1 demonstrates a schematic presentation of PTSO_TL for thalassemia detection.

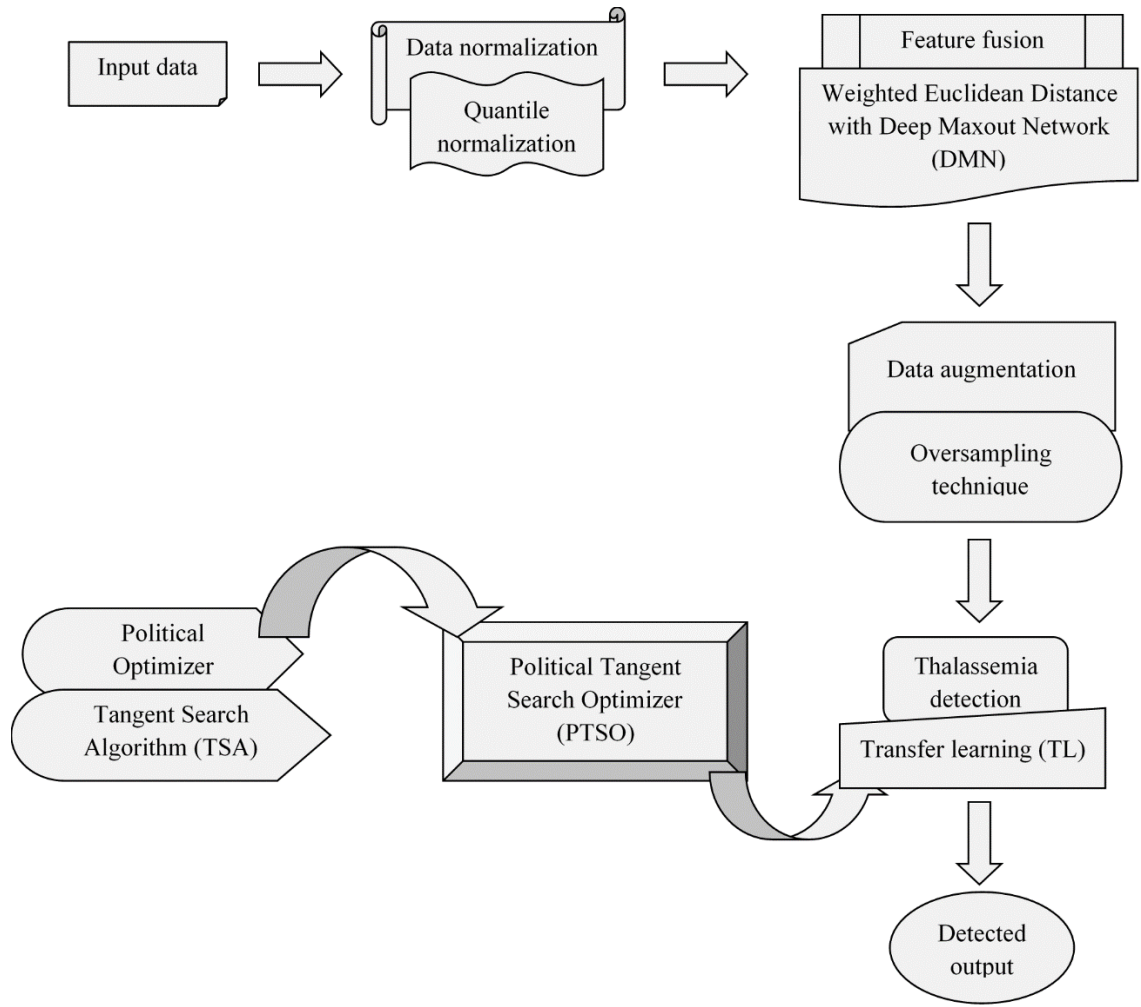

**Figure 1.** Schematic presentation of PTSO_TL for thalassemia detection

## Data acquisition

Assuming input data to detect thalassemia acquired from the dataset mentioned in [23] to be modeled by:

$$Q = \{Q_1, Q_2, ... Q_t ... Q_a\} \quad (1)$$

Here, total data present in the dataset $Q$ is represented as $Q_a$, whereas $t^{th}$ input data is indicated by $Q_t$.

## Data normalization utilizing quantile normalization

The goal of data normalization is to remove or lessen technical unpredictability. The common strategy of several normalization methods is the re-distribution of signal intensities over every sample, and thus all have a similar distribution. Here, quantile normalization is utilized to execute the data normalization process by considering input as $Q_t$.

Quantile normalization [16] is a significant normalization method that is normally employed in higher-dimensional data assessment. The processing of quantile normalization is easier, and it initially involves gene ranking of individual samples by means of magnitude, evaluating average values for the genes to occupy similar ranks, and thereafter substituting values of every gene to occupy specific ranks along with average values. The following step is reordering individual samples' genes in actual order. These sequences of steps describe quantile normalization, and the fundamental process underneath several strategies is explicated as follows:

Among quantile normalization strategies, 'All' normalizes the data as a single entire set, irrespective of classes and batch factors. The "Class-specific" strategy divides data by using phenotype classes first, wherein classes are quantile normalized separately, and normalized splitting is reintegrated to a single dataset. The "Discrete" strategy obtains the "Class-specific" method additionally and accounts for batch factor. Individual splitting is thereafter quantile normalized separately and reintegrated into a single database. The "Ratio" strategy involves the generation of a ratio matrix acquired by randomly comparing the samples from a single class against other samples that belong to another class. The normalized data obtained from quantile normalization is signified by $N_t$ with dimension $b \times c$.

## Feature fusion

Feature fusion assists in learning about every feature entirely, and the information on rich internal aspects thus helps in enhancing thalassemia detection. Here, Weighted Euclidean Distance with DMN is employed to execute feature fusion, and the normalized data $N_t$ with dimension $b \times c$ is given as input.

### Sort the feature based on Weighted Euclidean Distance

Euclidean distance is a highly helpful approach that is utilized in a wide variety of applications. Frequently, utilizing normalized Euclidean distance is preferred as it provides the average outcome of each distance. This kind of distance is called weighted Euclidean distance [17] while preferring to provide diverse significance to each distance rather than providing similar significance. The formulation to compute weighted Euclidean distance is shown as:

$$E_t = \sqrt{\sum_t W_t (y_t - z_t)^2} \qquad (2)$$

Where, $W_t = y_t$, if $y_i \neq 0$ and otherwise, $W_t = 1$ whereas $y$ implies candidate feature and $z$ specifies target.

### Fusion

The high-dimension feature fusion is developed with image and medical features. The fused features have complete details of patients. Thus the features are learned fully and help in thalassemia detection. The fusion formula can be illustrated by:

$$\Im^{new} = \sum_{t=1}^{a} \frac{\alpha}{\xi} \Im_t \qquad 1 \leq \xi \leq a \qquad (3)$$

$$t = t + \frac{Z}{i} \qquad (4)$$

$$i = \frac{Z}{a} \qquad (5)$$

Where, $Z$ it implies a count of features and $a$ represents the features to be selected.

**Generating $\alpha$ using DMN**

The $\alpha$ value in Eq. (3) is generated utilizing DMN, and the ground truth $\alpha$ can be represented as follows:

$$\alpha = E_t(d_t, \lambda_t) \qquad (6)$$

Here, $d_t$ implies data record whereas $\lambda_t$ denotes the average of $d_t$ belonging to the class, and an input given to DMN is the count of features. The generation matrix is delineated in Figure 2, and the architecture of DMN is elucidated in the below sub-section.

**Figure 2.** Generation matrix

### (a) Architecture of DMN

The system of maxout is a feed-forward formation such as multilayer perceptrons or deep CNN that employs a new type of activation function known as maxout unit. The DMN [18] consists of various layers, which develop hidden activations by means of maxout function.

*(i) Rectified linear unit (ReLU):* At first, ReLU is employed in Restricted Boltzmann Machines (RBM), which is illustrated by:

$$k_\gamma = \begin{cases} Y_\gamma, & if\ Y_\gamma \geq 0 \\ 0, & if\ Y_\gamma < 0 \end{cases} \qquad (7)$$

Here, $Y_\gamma$ indicates input given to neuron, whereas $k_\gamma$ signifies output.

*(ii) Maxout:* Maxout is a common kind of the ReLU that obtains the max function on $v\,(v=2)$ trainable linear operations. For input $Y \in Z^\tau$, wherein $Y$ denotes state vector or a raw input vector of hidden layer and maxout unit output is represented by:

$$N_\gamma(Y) = \max_{\omega \in [1,v]} I_{\gamma\omega} \tag{8}$$

Here, $I_{\gamma\omega} = Y^E A_{\ldots\gamma\omega} + V_{\gamma\omega}$, $A \in Z^{\tau \times \rho \times v}$ implies trainable parameters, and the count of linear sub-hidden units is denoted by $v$. In CNN, the maxout unit activation equalizes the maximum over $v$ feature maps. Moreover, maxout unit is similar to normally utilizing spatial maxpooling in the structure of CNN. It acquires a high value over identical input, wherein the spatial maxpooling is linked to $v$ diverse input.

*(iii) DMN:* It is a form of trainable activation operation together with numerous layer configurations. For an input $Y \in Z^\tau$, wherein $Y$ denotes state vector or a raw input vector of hidden layer, the activation of a hidden unit is computed by:

$$H^1_{\gamma,\omega} = \max_{\omega \in [1,v_1]} Y^E A_{\ldots\gamma\omega} + V_{\gamma\omega} \tag{9}$$

$$H^2_{\gamma,\omega} = \max_{\omega \in [1,v_2]} {H^1_{\gamma,\omega}}^E A_{\ldots\gamma\omega} + V_{\gamma\omega} \tag{10}$$

$$H^\rho_{\gamma,\omega} = \max_{\omega \in [1,v_\rho]} {H^{\rho-1}_{\gamma,\omega}}^E A_{\ldots\gamma\omega} + V_{\gamma\omega} \tag{11}$$

$$H^n_{\gamma,\omega} = \max_{\omega \in [1,v_n]} {H^{n-1}_{\gamma,\omega}}^E A_{\ldots\gamma\omega} + V_{\gamma\omega} \tag{12}$$

$$N_\gamma = \max_{\omega \in [1,v_n]} H^n_{\gamma,\omega} \tag{13}$$

Where, $v_\rho$ indicates overall units in $\rho^{th}$ layer and $n$ represents whole layers in DMN. The DMN activation is stronger in estimating an arbitrary recurrent activation operation other than the non-convex activation functions. Traditional non-linear activation operations such as rectified linear as well as absolute value rectifiers are evaluated by DMN when $v$ is not less than 2. Even though an efficient extraction of a feature requires a much more complicated non-linear function, the DMN is able to evaluate an arbitrary activation function through augmenting parameter $v$. The fused output is symbolized by $M_t$ with dimension $b \times e$, where $c > e$ and DMN architecture is shown in Figure 3.

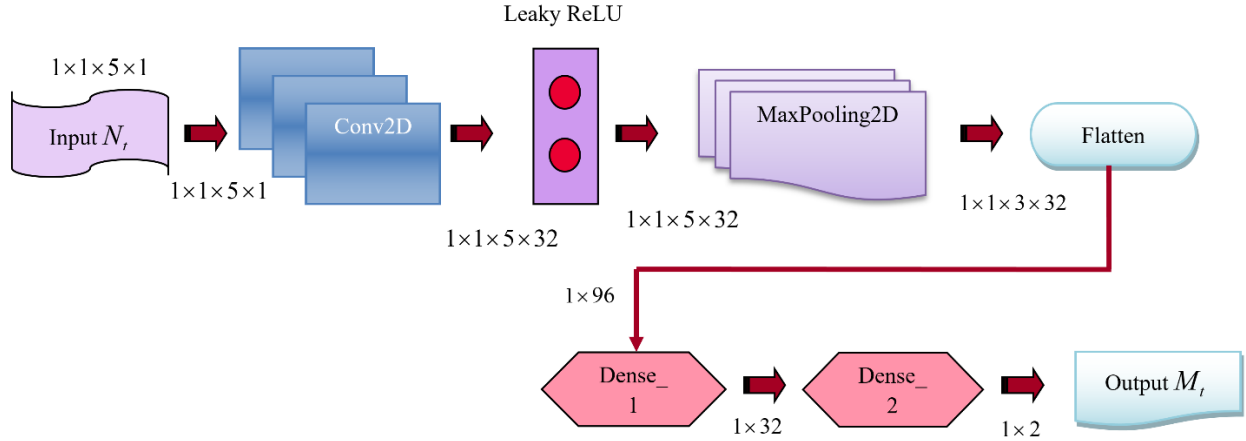

**Figure 3.** Architecture of DMN

## Data augmentation utilizing oversampling method

Data augmentation performs as the regularizer in avoiding overfitting in NN and enhances performance in imbalanced class issues. Here the oversampling technique is utilized to carry out data augmentation considering $M_t$ with dimension $b \times e$.

The problem of class imbalance has consisted of several data monitoring malfunctions that include failure description text. Various samples are contained in major-class while compared to minor-class. Essentially, the major-class occurs repeatedly but usually performs less damage, while the minor-class occurs rarely, but it does much damage. This kind of unbalanced class data is a vital challenge for performing classification.

At present, data balancing processing of oversampling [29] takes place in two modes. In one mode the samples of minor-class are copied directly, whereas in the other samples of minor-class are artificially created accordingly to minor-class features. An earlier is simple to carry out, even though it causes the over-fitting, whilst the final is much more complex but harder to over-fit.

Assume a group of samples $X = \{x_o | o = 1,2,3,...,\varpi\}$ of a minor-class. At some point in $X$, Euclidean distance is computed amongst these points and all remaining points to attain $\partial$ nearby points. Consider oversampling multiplier specified by $\zeta$ that chooses randomly at $\zeta$ points in $\partial$ nearby points to generate $A = \{\hat{x} | \phi = 1,2,3,...,\zeta\}$. The newer samples are added to $X$ by random linear interpolation and are given by:

$$x_{new}(o,\phi) = \{x_o + rand(0,1)(\hat{x}_\phi - x_\phi) | o = 1,2,3,...,\varpi; \phi = 1,2,3,...,\zeta\} \quad (14)$$

Where, $rand(0,1)$ implies a random number between 0 and 1. The aforementioned formula produces $\varpi$ samples of a minor-class to obtain the database balancing goal. The augmented data is illustrated by $O_t$ with dimension $n \times e$, where $b < n$.

## Thalassemia detection utilizing TL

Thalassemia-affected patients have diverse degrees of lower red blood cell (RBC) values and enlargement of the spleen and liver, depending upon a kind of inherited defects in RBC hemoglobin generation. Here, CNN with TL is employed for the detection of thalassemia, wherein CNN is used with hyperparameters from trained models such as Xception. PTSO is utilized for training CNN with TL, where PTSO is an incorporation of PO with TSA. The input given to execute thalassemia detection is $O_t$ with dimension $n \times e$.

### Architecture of TL

CNN is the function for mapping input data to output and is normally comprised of layers like convolutional, max, or average pooling, activation layers such as sigmoid or Rectified Linear Unit (ReLU), and softmax for detection of output. For input data, normally, the initial convolutional layer extracts the edges, and the following convolutional layer operates as a high-level feature extractor. The pooling layer is utilized for computing the maximal or average of individual patches of the feature map from the prior layer. The role of the activation layer is to identify saturation issues. The CNN is mathematically denoted as $L$ and can be illustrated by:

$$H = h_Y h_{Y-1} h_{Y-2} h \ \ldots\ldots \ \ldots..h_2 h_1 \tag{15}$$

An individual function illustrates the layer that obtains the output of the prior layer.

After training, CNN is utilized as a classifier or a feature extractor in the event of TL [19] [35]. The TL is a method utilized for improving machine learning performance by harnessing information acquired by other tasks. Here, CNN is utilized with hyperparameters from the trained models like Xception.

Xception [20] is a deep CNN structure that is constructed in relation to the inception modules, wherein inception modules are substituted by depth-wise separable convolutions. Xception performs better while investigated in a larger dataset. Xception V3 assists in offering the most effective utilization of hyperparameters. The depth-wise distinguishable convolutions are executed with linear as well as residual associations. This architecture comprises of 36 convolutional layers that form the feature extraction basis of the network. The CNN used with hyperparameters of trained Xception is shown in Figure 4, and thalassemia detected output is signified by $T_t$:

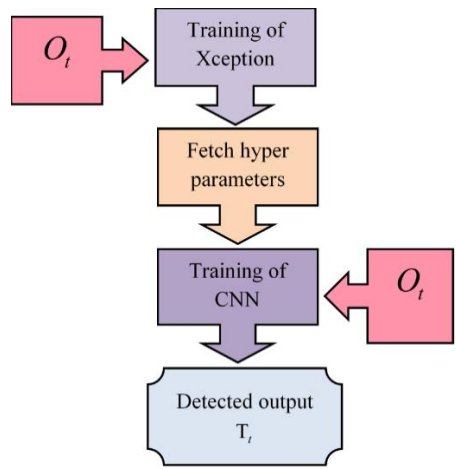

**Figure 4**. Process of CNN used with hyperparameters of trained Xception

**Training of TL utilizing PTSO**

Development of the PO [22] is driven by the multiple staged political processes, and it refers to the mathematical mapping of every main stage of politics, like constituent allocation, switching of parties, inter-party elections, parliament affairs, and election campaigns. It is invariant for functioning shifting and constantly executes in higher dimensional search spaces. TSA [21] is a population-based algorithm that is designed to resolve optimization issues. It utilizes a mathematical system on the basis of tangent function to progress a provided solution towards the finest solution. TSA is simple and effective as well as offering a lesser count of parameters. Here, a combination of PO with TSA specified as PTSO is utilized for training TL for the detection of thalassemia.

**Tangent search position encoding**

The learning parameter $\psi$ is tuned in a search space $\hat{\lambda}$ to acquire an optimum solution, and it is given by $\hat{\lambda} = [1 \times \psi]$.

**Fitness function**

The variation is taken among TL and targeted outputs to compute fitness measure and is modeled by:

$$\Psi = \frac{1}{a} \sum_{t=1}^{a} [C_t - T_t]^2 \qquad (16)$$

In the above equation $a$ indicates total data, $C_t$ and $T_t$ represent target output and output obtained from TL.

PTSO is utilized to train TL to perform the below-mentioned steps and acquire a better solution.

*Step 1: Initialize the solution*

TSA begins by creating an initial population randomly within solution space boundaries. An initial solution is distributed uniformly all over search space that can be represented by:

$$T = \{T_1, T_2, ..., T_m, ..., T_l\} \qquad (17)$$

Here, $T_m$ implies $m^{th}$ a candidate solution and $l$ specifies the count of variables, and $T$ signifies population.

*Step 2: Objective function evaluation*

The evaluation of the objective function is done by finding differences between TL and targeted outputs utilizing Eq. (16).

*Step 3: Intensification search*

In an intensification search, a randomly local walk is directed initially, and then a few variables of the attained solution are substituted by values of the related variable in the present optimum solution.

$$T_m^{q+1} = T_m^q + s * \tan(\theta) * (T_m^q - optU_m^q) \qquad (18)$$

$$T_m^{q+1} = T_m^q + s * \tan(\theta) * T_m^q - s * \tan(\theta) optU_m^q \quad (19)$$

$$T_m^{q+1} = T_m^q [1 + s * \tan(\theta)] - s * \tan(\theta) optU_m^q \quad (20)$$

The standard equation of PO is given by:

$$J_{u,w}^v(q+1) = J_{u,w}^v(q-1) + R(J_{u,w}^v(q) - J_{u,w}^v(q-1)), \quad if \ J_{u,w}^v(q-1) \leq \mu^* \leq J_{u,w}^v(q) \ (or) \ J_{u,w}^v(q-1) \geq \mu^* \geq J_{u,w}^v(q) \quad (21)$$

Where,

$$J_{u,w}^v(q+1) = T_m^{q+1} \quad (22)$$

$$J_{u,w}^v(q-1) = T_m^{q-1} \quad (23)$$

$$J_{u,w}^v(q) = T_m^q \quad (24)$$

Then Eq. (21) becomes:

$$T_m^{q+1} = T_m^{q-1} + R(T_m^q - T_m^{q-1}) \quad (25)$$

$$T_m^q = \frac{T_m^{q+1} - T_m^{q-1}(1-R)}{R} \quad (26)$$

Substitute Eq. (26) in Eq. (20), then equation becomes:

$$T_m^{q+1} = \left(\frac{T_m^{q+1} - T_m^{q-1}(1-R)}{R}\right)[1 + s * \tan(\theta)] - s * \tan(\theta) optU_m^q \quad (27)$$

$$T_m^{q+1} - \frac{T_m^{q+1}}{R}[1 + s * \tan(\theta)] = \frac{T_m^{q-1}(R-1)[1 + s * \tan(\theta)] + R.s * \tan(\theta) optU_m^q}{R} \quad (28)$$

$$\frac{(R - 1 - s * \tan(\theta))T_m^{q+1}}{R} = \frac{T_m^{q-1}(R-1)[1 + s * \tan(\theta)] + R.s * \tan(\theta) optU_m^q}{R} \quad (29)$$

The updated equation of PTSO is modeled by:

$$T_m^{q+1} = \frac{T_m^{q-1}(R-1)[1 + s * \tan(\theta)] + R.s * \tan(\theta) optU_m^q}{R - 1 - s * \tan(\theta)} \quad (30)$$

Here, $R$ denotes random number between 0 and 1 whereas $T_m^q$ denotes $m^{th}$ solution at $q^{th}$ iteration and $s$ implies step size.

When few variables of the attained solution are replaced by related variable, values in the optimum solution are specified by:

$$T_m^{q+1} = optU_m^q, \quad \text{if variable } m \text{ is chosen} \tag{31}$$

Accordingly, an acquired solution $T_m^{q+1}$ has a similarity proportion below 50%, along with a better present solution that assists in improving the present solution locally.

An individual acquired solution $T$ is restored by the below equations that address the problem of values such as lower bound $\ell$ and upper bound $r$.

$$\left.\begin{array}{l} T(T < \ell) = \Re * (r - \ell) + \ell \\ T(T > \ell) = \Re * (r - \ell) + \ell \end{array}\right\} \tag{32}$$

### Step 4: Exploration search

In contrast to local search techniques, population-based metaheuristic-enabled algorithms have a high exploration ability. TSA utilizes the product of the variable step size and tangent flight for making globally random walks. A tangent operation assists in exploring effectively in search space, certainly $\theta$ nearer to $\pi/2$ makes the tangent function value larger and the attained solution is far away from the present solution, whereas $\theta$ nearer to 0 provides smaller tangent function values and the attained solution is closer to the present solution. An equation of exploration search is applied on individual variables along with probability similar to $1/M$, where $M$ indicates problem dimension:

$$T_m^{q+1} = T_m^q + s * \tan(\theta) \tag{33}$$

An intensification, as well as exploration search, is applied in accordance with provided probability, known as $\text{P}switch$.

### Step 5: Escape the minimum local process

To escape from local minimal stagnation issues, TSA integrates a method that utilizes a certain process. A process has two parts accomplished with few probability $\text{P}esc$. In an individual iteration, a single search agent is chosen in a random manner, and the below equation is applied. Furthermore, the newer random solution is able to replace the worst solution with a probability of 0.01.

$$T = T + K * (optU - R * (optU - T)) \tag{34}$$

$$T = T + \tan(\theta) * (r - \ell) \tag{35}$$

### Step 6: Termination

PTSO obtains a better solution by performing the steps mentioned above, and algorithm 1 explicates the pseudo code of PTSO.

**Algorithm 1**. Pseudo code of PTSO

| SL. No | Pseudo code of PTSO |
|---|---|
| 1 | **Input:** $T_m^q, s, r, \ell$ |
| 2 | **Output:** $T_m^{q+1}$ |

| 3 | **Begin** |
|---|---|
| 4 | Initialize the population |
| 5 | Computation of fitness value using Eq. (16) |
| 6 | ***while*** $q <=$ maximal function estimations |
| 7 | ***for*** individual search agent $T_m$ |
| 8 | ***if*** $R < \mathrm{P}switch$ |
| 9 | Apply an intensification search utilizing Eq. (30) |
| 10 | ***else*** |
| 11 | Apply an exploration search utilizing Eq. (33) |
| 12 | ***end*** |
| 13 | ***end for*** each |
| 14 | ***if*** $R < \mathrm{P}esc$ |
| 15 | Select the search agent randomly |
| 16 | Apply the escape local minima utilizing Eq. (34) |
| 17 | $q = q + 1$ |
| 18 | ***end while*** |
| 19 | **Terminate** |

# Results and Discussion

PTSO_TL achieved the finest outcomes for thalassemia detection, and the assessments carried out are illustrated in this portion.

## Experiment Setup

Experimentation of PTSO_TL designed for thalassemia detection is accomplished in the PYTHON tool.

## Dataset Description

The Alpha Thalassemia Dataset [23] consists of 288 cases that were gathered from Alpha Thalassemia carrier children as well as their family members screened from 2016 to 2020. It has 15 independent variables, and in addition to sex ('female' and 'male'), all the others are continual variables in a float format.

## Performance Measures

Precision, recall, and f-measure are the considered performance measures to evaluate PTSO_TL for thalassemia detection.

## Precision

Precision evaluates the detected positive cases of thalassemia and can formulate utilizing the below equation:

$$\aleph_1 = \frac{\hbar}{\hbar + \varsigma} \tag{36}$$

Here, $\hbar$ implies true positive whereas $\varsigma$ specifies false positive.

**Recall**

Recall refers to the fraction of truly positive cases out of overall positive cases, and it is computed as follows:

$$\aleph_2 = \frac{\hbar}{\hbar + \eta} \tag{37}$$

Where, $\eta$ denotes false negative.

**F-Measure**

F-measure is defined as harmonic mean amongst precision and recall, which is calculated as given below:

$$\aleph_3 = 2 \div \frac{\aleph_2 \times \aleph_1}{\aleph_2 + \aleph_1} \tag{38}$$

## Performance Analysis

Figure 5 presents a performance evaluation of PTSO_TL by varying learning sets with several iterations. The estimation of PTSO_TL considering precision is shown in Figure 5 a). The precision obtained by PTSO_TL is 0.773, 0.806, 0.834, 0.874, and 0.931 with iterations 10, 20, 30, 40, and 50 for learning set=90%. Figure 5 b) illustrates the assessment of PTSO_TL as regards recall. Recall attained by PTSO_TL is 0.809 with iteration 10, 0.829 with iteration 20, 0.850 with iteration 30, 0.882 with iteration 40, and 0.952 with iteration 50 when learning set=90%. Analysis of PTSO_TL concerning f-measure is delineated in Figure 5 c). Recall acquired by PTSO_TL is 0.791, 0.817, 0.842, 0.878, and 0.942 with iterations 10, 20, 30, 40 and 50 while learning set=90%.

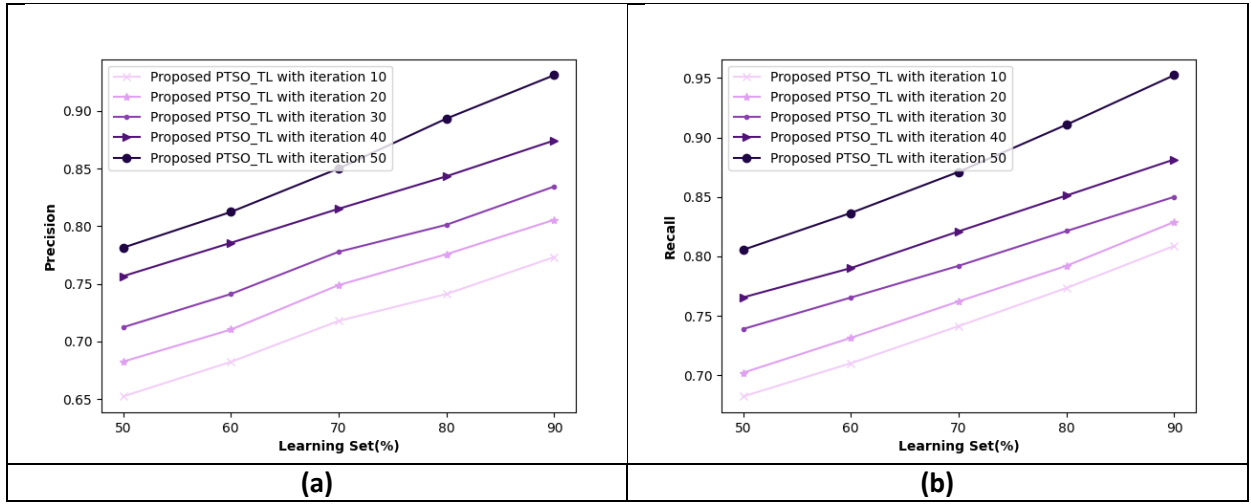

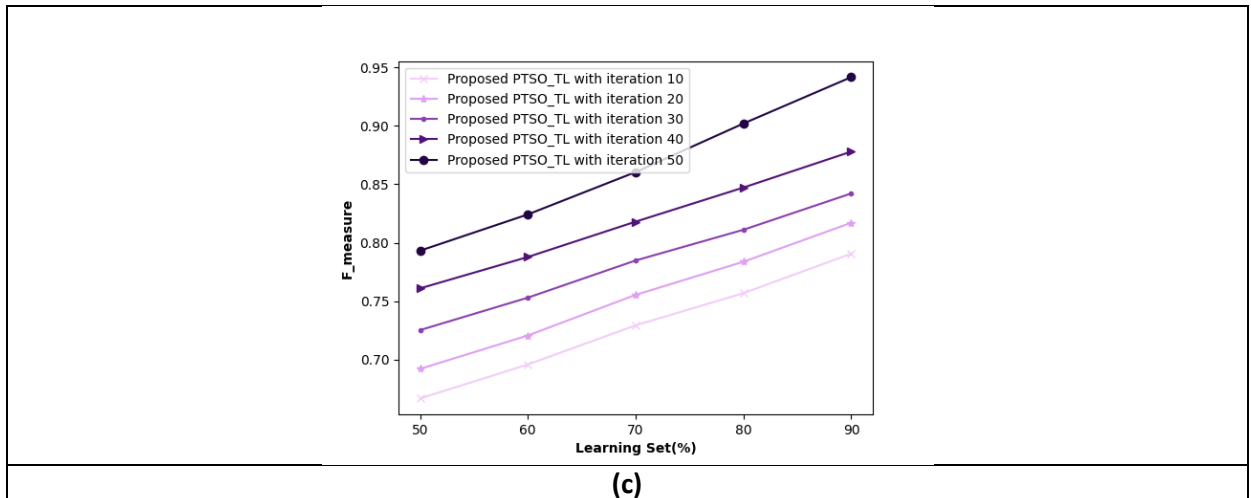

(c)

**Figure 5.** Performance estimation of PTSO_TL, a) Precision, b) Recall, c) F-measure

## Comparative Techniques

SVM [1], Supervised ML [3], ThalPred model [4], and ML [5] are the existing techniques used for comparative assessment to reveal the efficacy of PTSO_TL.

## Comparative Analysis

PTSO_TL is compared with existing approaches to perform a comparative assessment by varying the learning set and k value.

### Analysis Based Upon Learning Set

Comparative estimation of PTSO_TL is done by using measures for varying the learning sets, as explicated in Figure 6. Figure 6 a) expounds analysis of PTSO_TL based on precision. PTSO_TL obtained a precision of 0.935 whereas SVM, Supervised ML, ThalPred model, and ML achieved 0.741, 0.788, 0.831, and 0.875, revealing performance enhancement of 20.756%, 15.732%, 11.088% and 6.352% for learning set=90%. The estimation of PTSO_TL with regard to recall is displayed in Figure 6 b). The recall value acquired by PTSO_TL is 0.958 while learning set=90%, whereas recall achieved by SVM is 0.811, Supervised ML is 0.831, ThalPred model is 0.866, and ML is 0.903. Thus, the performance of PTSO_TL is enhanced by 15.306%, 13.215%, 9.618%, and 5.671%. Figure 6 c) shows an assessment of PTSO_TL concerning f-measure. When learning set=90%, PTSO_TL acquired an f-measure of 0.946 whereas SVM, Supervised ML, ThalPred model, and ML attained 0.774, 0.809, 0.848, and 0.889, which showed performance improvement of about 18.155%, 14.507%, 10.368%, and 6.017%, respectively.

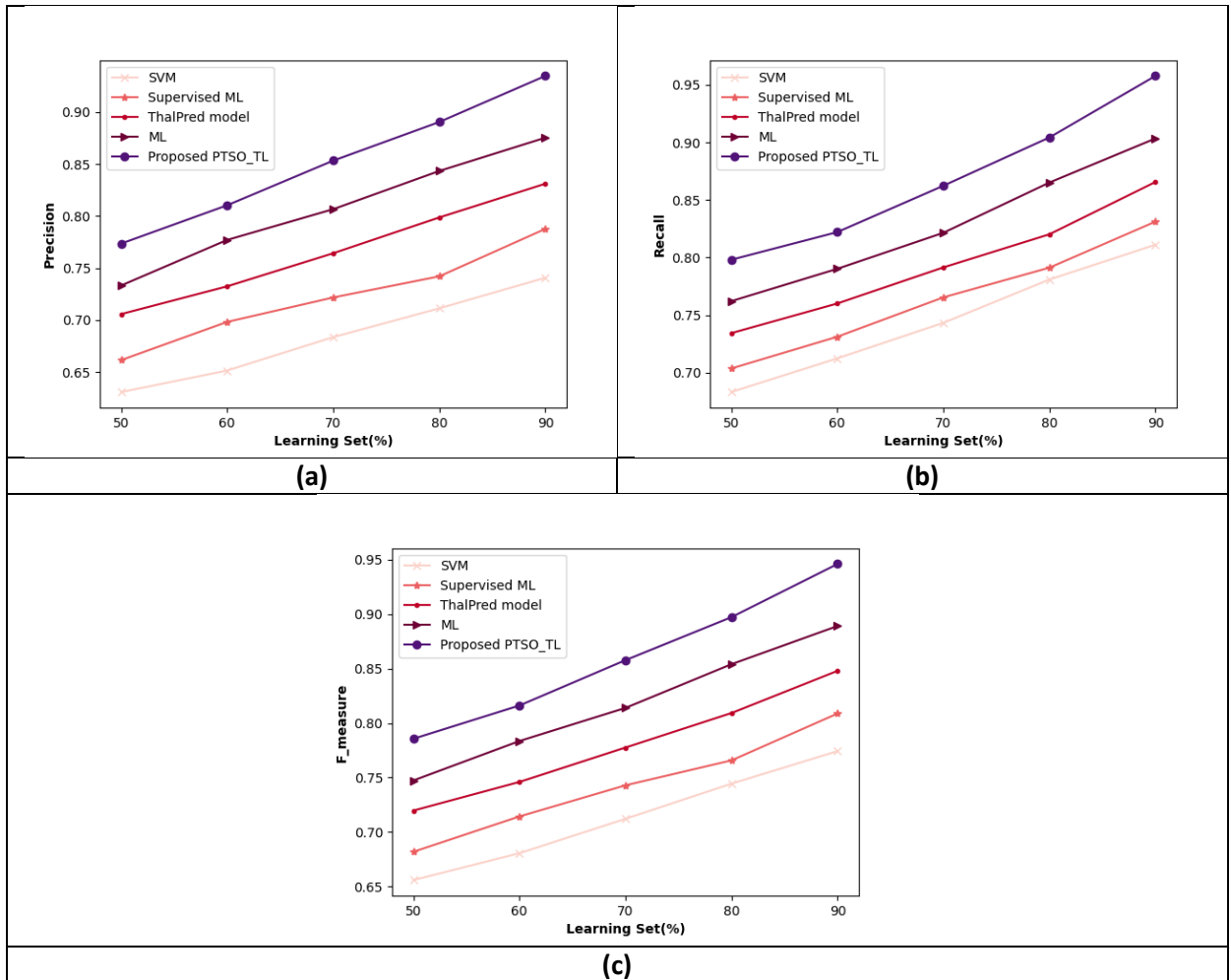

**Figure 6.** Comparative estimation of PTSO_TL based on learning set: a) Precision, b) Recall, c) F-measure

**Analysis Based Upon K Value**

Figure 7 elucidates the assessment of PTSO_TL using measures for changing the k value. Figure 7 a) demonstrates the evaluation of PTSO_TL considering precision. PTSO_TL achieved a precision of 0.943 whereas the precision obtained by SVM is 0.769, Supervised ML is 0.802, ThalPred model is 0.847, and ML is 0.890 when k value=9. It shows the enhanced performance of PTSO_TL of about 18.489%, 14.968%, 10.251%, and 5.613%. Analysis of PTSO_TL with respect to recall is interpreted in Figure 7 b). The value of recall acquired by PTSO_TL is 0.961 while k value=9 whereas SVM, Supervised ML, ThalPred model, and ML attained 0.823, 0.856, 0.889, and 0.928. This describes performance enhancement of PTSO_TL of about 14.323%, 10.983%, 7.515%, and 3.479%. Figure 7 c) explicates the analysis of PTSO_TL considering the f-measure. When k value=9, PTSO_TL achieved an f-measure of 0.952 whereas the f-measure value obtained by SVM is 0.795, Supervised ML is 0.828, ThalPred model is 0.867 and ML is 0.909, revealing improvement in the performance of PTSO_TL of about 16.478%, 13.040%, 8.916%, and 4.568%, respectively.

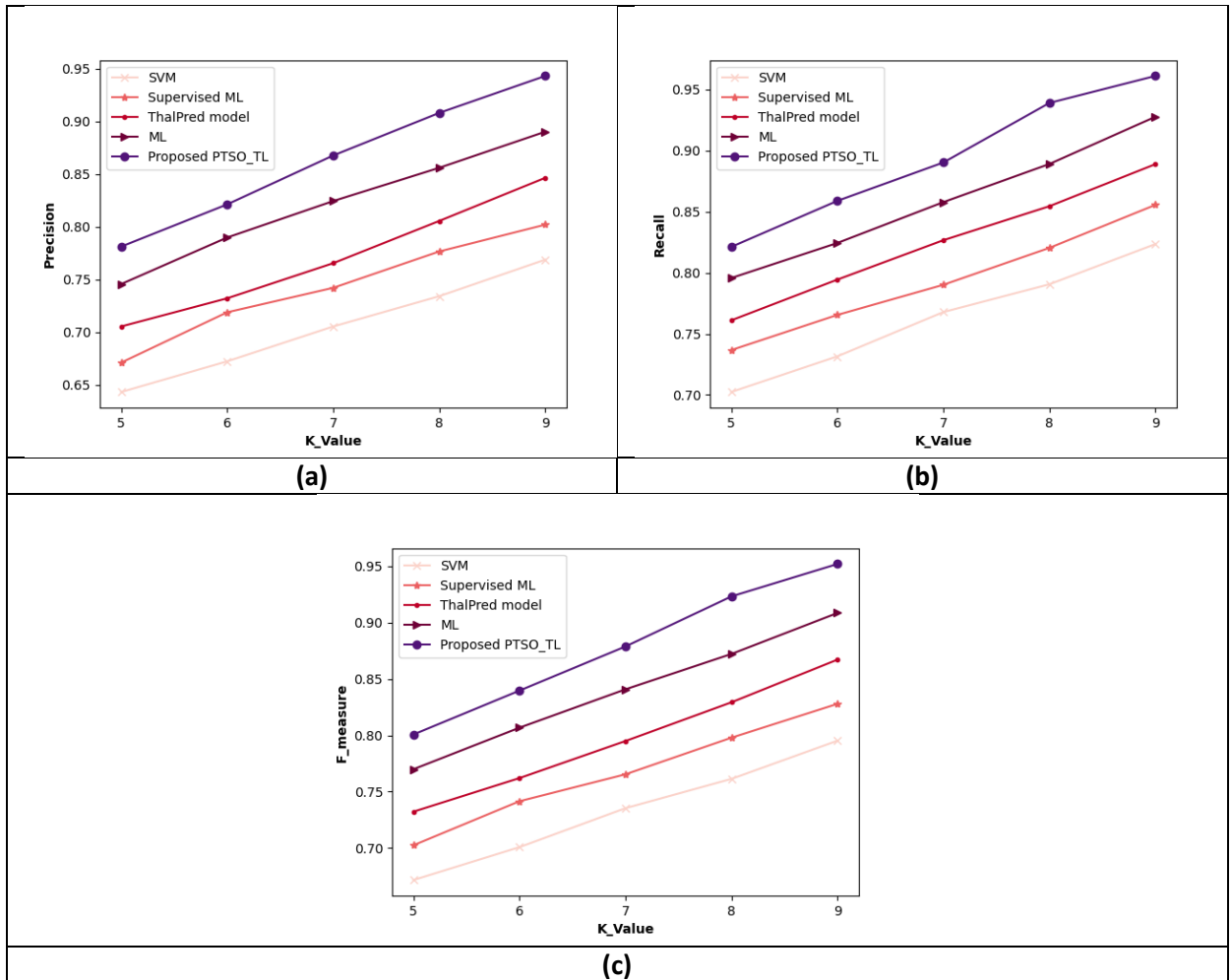

**Figure 7.** Comparative estimation of PTSO_TL based on k value, a) Precision, b) Recall, c) F-measure

## Algorithmic Analysis

Rat Swarm Optimizer (RSO) [28]+TL, Honey Badger Algorithm (HBA) [27]+TL, TSA [21]+TL, and PO [22]+TL are the algorithms compared with PTSO+TL to show its effectiveness.

Assessment of PTSO+TL by varying iterations with consideration of measures is demonstrated in Figure 8. Figure 8 a) interprets the evaluation of PTSO+TL on the basis of precision. PTSO+TL acquired a precision value of 0.931, whereas RSO+TL, HBA+TL, TSA +TL, and PO+TL attained 0.804, 0.847, 0.871, and 0.911. This reveals performance improvement of PTSO+TL by 13.629%, 9.091%, 6.433%, and 2.173% while iteration=50. Analysis of PTSO+TL considering recall is explicated in Figure 8 b). When iteration=50, recall attained by PTSO+TL is 0.952, whereas recall obtained by RSO+TL is 0.815, HBA+TL is 0.848, TSA +TL is 0.889, and PO+TL is 0.921. The performance of PTSO+TL is improved by 14.346%, 10.945%, 6.630%, and 3.245%. Figure 8 c) delineates the estimation of PTSO+TL with regard to the f-measure. PTSO+TL attained an f-measure of 0.942, whereas RSO+TL, HBA+TL, TSA +TL, and PO+TL acquired 0.810, 0.847, 0.880, and 0.916 for iteration=50. Enhancement of the performance of PTSO+TL based on iterations is therefore about 13.985%, 10.018%, 6.531%, and 2.706 %.

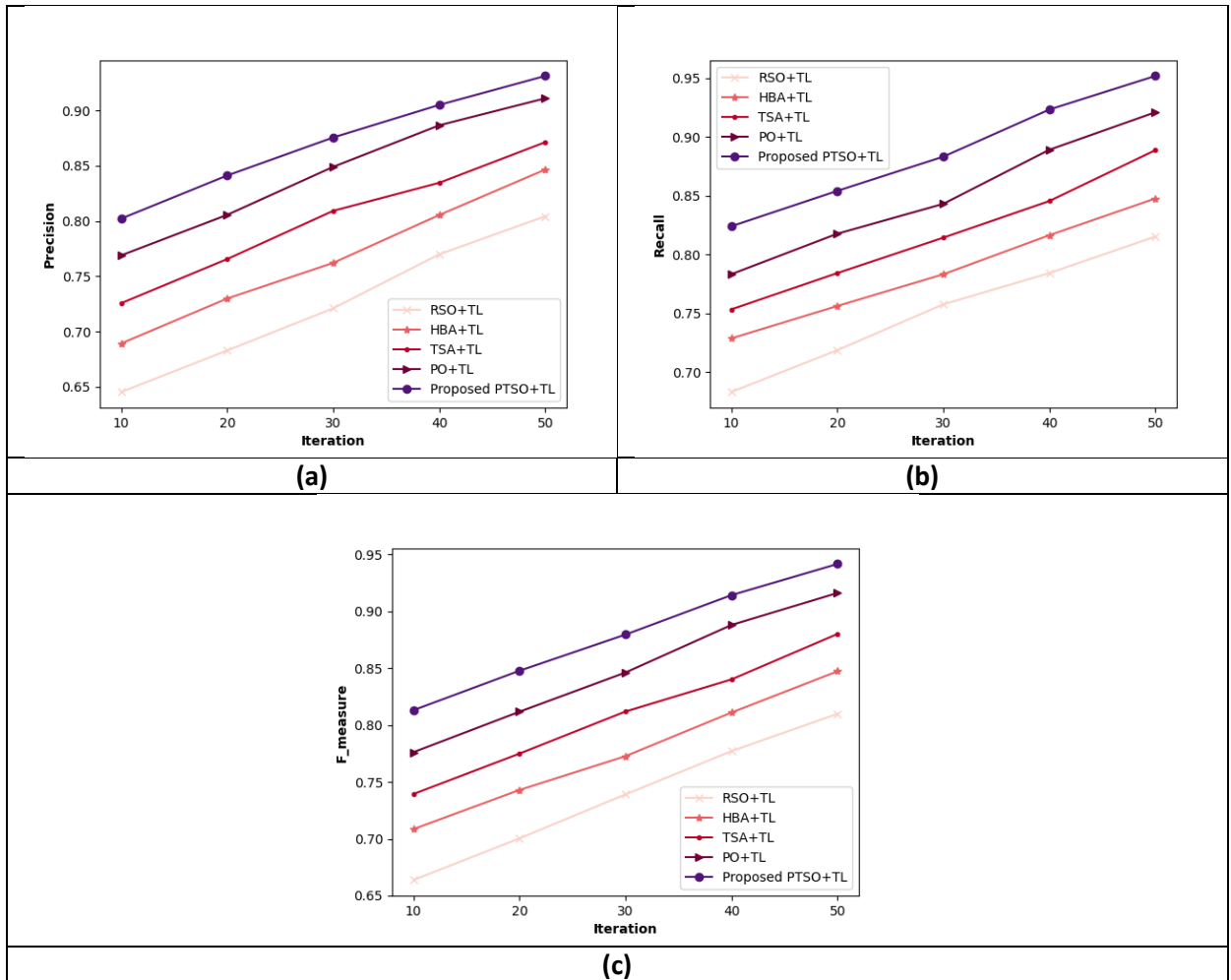

**Figure 8.** Algorithmic analysis of PTSO+TL, a) Precision, b) Recall, c) F-measure

## Comparative Discussion

PTSO_TL obtained good outcomes in comparison with SVM, Supervised ML, ThalPred model, and ML, as described in Table 1. From the discussion table, it can be seen that PTSO_TL obtained maximal precision, recall, and f-measure values of 94.3%, 96.1%, and 95.2%, when the k value is considered as 9. Thus, the results attained by PTSO_TL have shown its effectiveness for thalassemia detection.

**Table 1.** Comparative discussion of PTSO+TL

| Analysis based on | Metrics/Methods | SVM | Supervised ML | ThalPred model | ML | Proposed PTSO_TL |
|---|---|---|---|---|---|---|
| **Learning set=90%** | Precision | 74.1% | 78.8% | 83.1% | 87.5% | 93.5% |
|  | Recall | 81.1% | 83.1% | 86.6% | 90.3% | 95.8% |
|  | F-measure | 77.4% | 80.9% | 84.8% | 88.9% | 94.6% |
| **K value=9** | Precision | 76.9% | 80.2% | 84.7% | 89.0% | **94.3%** |

|  | Recall | 82.3% | 85.6% | 88.9% | 92.8% | **96.1%** |
|--|--------|-------|-------|-------|-------|-----------|
|  | F-measure | 79.5% | 82.8% | 86.7% | 90.9% | **95.2%** |

# Conclusion

Thalassemia is a global hereditary disorder and a main public health issue resulting in an abnormal ratio of hemoglobin sub-units. The technique to deal with the thalassemia problem is to avoid or manage the birth of new cases. This requires the precise detection of couples at higher threat of thalassemia. Recently, several methods have been developed for detecting thalassemia. In this research, PTSO_TL is presented for the detection of thalassemia. The input data are obtained from a specific database. Quantile normalization is utilized for normalizing data in the data normalization stage. Thereafter, feature fusion is performed using Weighted Euclidean Distance with DMN. Afterward, fused data is augmented in the data augmentation phase utilizing the oversampling method. Finally, thalassemia detection is executed utilizing CNN with TL, wherein CNN is used with hyperparameters from trained models such as Xception. The CNN with TL is trained to employ PTSO, which is the integration of PO with TSA. The PTSO_TL acquired maximum precision of 94.3%, recall of 96.1%, and f-measure of 95.2% when considering k value=9. In the future, this work will be extended by utilizing image datasets from other blood-related diseases.

### Data availability

The data supporting the findings of this study are available from the corresponding author upon reasonable request.